\documentclass[letterpaper, 10 pt, conference]{ieeeconf}  %

\IEEEoverridecommandlockouts                              %

\usepackage{subcaption}
\usepackage{graphicx}
\usepackage{tikz}
\usepackage{comment}
\usepackage{amsmath,amssymb} %
\usepackage{color}
\usepackage{colortbl}
\usepackage{multicol}
\usepackage{multirow}
\usepackage{cite}
\usepackage{lipsum}
\usepackage{booktabs}
\usepackage{url}
\usepackage{hyperref}

\newcommand{\ie}{\textit{i}.\textit{e}., }
\newcommand{\eg}{\textit{e}.\textit{g}. }

\title{\LARGE \bf
Stereo-Inertial Poser: Towards Metric-Accurate Shape-Aware Motion Capture Using Sparse IMUs and a Single Stereo Camera
}

\author{Tutian Tang$^{12*}$, Xingyu Ji$^{12*}$, Yutong Li$^{1}$, MingHao Liu$^{1}$, Wenqiang Xu$^{1}$, and Cewu Lu$^{13}$%
\thanks{$^{1}$ School of Computer Science, Shanghai Jiao Tong University}%
\thanks{$^{2}$ Meta Robotics Institute, SJTU \; $^{3}$ Shanghai Innovation Institute}%
\thanks{
* Equal contribution.
This work was supported by the Science and Technology Major Project of Jiangsu Province (No. BG2024041),
by Shanghai Commitee of Science and Technology (No. 24511103200),
by the National Key Research and Development Project of China (No. 2022ZD0160102),
by Shanghai Artificial Intelligence Laboratory, XPLORER PRIZE grants,
and by Shanghai Municipal Education Commission (No. 2024AIYB010).
\tt\small \{tttang, ji\_xingyu, davidliyutong, lmh209, vinjohn, lucewu\} @sjtu.edu.cn}%
}

\begin{document}

\maketitle
\thispagestyle{empty}
\pagestyle{empty}

\begin{figure*}
    \centering
    \includegraphics[width=0.95\linewidth]{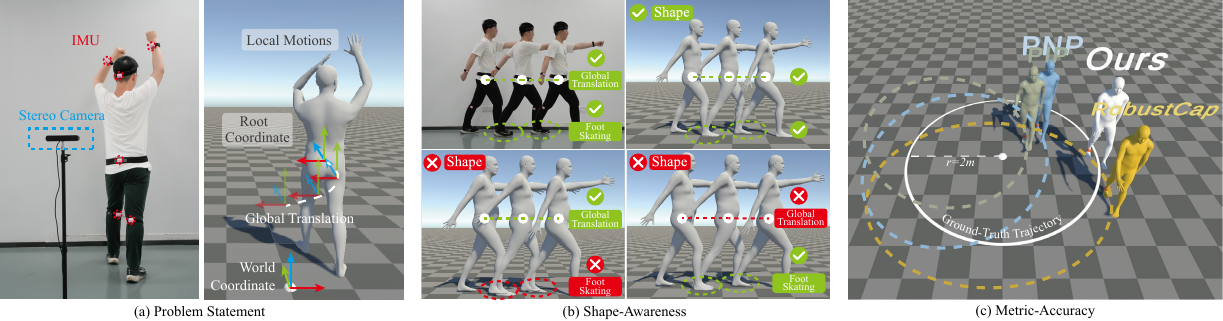}
    \caption{
    (a) Our method estimates full human motion, including the local motions in the body's root coordinate and the global translation, with six inexpensive wearable IMUs and a stereo camera.
    (b) Previous methods in this track usually ignore the \textit{body shape} (\ie $\boldsymbol{\beta}$ parameters of the SMPL model), which may lead to the foot-skating effect or the incorrect global translation estimation.
    (c) Qualitative comparison with previous methods~\cite{pip,pnp,robustcap}
    shows the proposed method can estimate metric-accurate, drift-free global translation in the world coordinate.
    }
    \label{fig:teaser}
\end{figure*}

\begin{abstract}
Recent advancements in visual-inertial motion capture systems have demonstrated the potential of combining monocular cameras with sparse inertial measurement units (IMUs) as cost-effective solutions, which effectively mitigate occlusion and drift issues inherent in single-modality systems. However, they are still limited by metric inaccuracies in global translations stemming from monocular depth ambiguity, and shape-agnostic local motion estimations that ignore anthropometric variations. We present Stereo-Inertial Poser, a real-time motion capture system that leverages a single stereo camera and six IMUs to estimate metric-accurate and shape-aware 3D human motion. By replacing the monocular RGB with stereo vision, our system resolves depth ambiguity through calibrated baseline geometry, enabling direct 3D keypoint extraction and body shape parameter estimation. IMU data and visual cues are fused for predicting drift-compensated joint positions and root movements, while a novel shape-aware fusion module dynamically harmonizes anthropometry variations with global translations. Our end-to-end pipeline achieves over 200 FPS without optimization-based post-processing, enabling real-time deployment. 
Quantitative evaluations across various datasets demonstrate state-of-the-art performance. Qualitative results show our method produces drift-free global translation under a long recording time and reduces foot-skating effects. The code, data, and supplementary materials are available at \url{https://sites.google.com/view/stereo-inertial-poser}.

\end{abstract}

\section{Introduction}
Human motion capture enables many robotic applications like human-robot interaction, teleoperation, and imitation learning~\cite{icra_mocap,icra_mocap2_loose}.
Current commercial solutions can be divided into optical-based, IMU-based, and hybrid ones according to the modality of sensors. They are precise but often expensive and confined to controlled environments, limiting their accessibility for real-world deployment. Therefore, developing a low-cost and easy-to-use motion capture system has been an active topic in the research community for a long time.
One straightforward strategy is to reduce the number of sensors, whether cameras or IMUs.
First, it is possible to estimate human motions from monocular videos~\cite{smplify,frankmocap,hybrik}, but they are vulnerable to occlusion, extreme light conditions, and depth ambiguities.
Second, pure inertial systems~\cite{sip,pip,pnp} suffer from the drifting issue due to sensor noise.
Third, to avoid the inherent limitations of single-modality systems, recent hybrid solutions introduce visual-inertial fusion, combining a single RGB camera with six sparse IMUs~\cite{robustcap,hybridcap}.
By decoupling the full human motion into \textit{global translations} in the world coordinate and \textit{local motions} in the body's root coordinate, they demonstrate reduced drifting effects and improved robustness against occlusion.

However, fundamental limitations still persist in current visual-inertial fusion systems. First, the predicted translations from pure 2D images are \textbf{not} \textit{metric-accurate}, due to the well-known depth ambiguity problem.
Second, the estimated local motions are \textbf{not} \textit{shape-aware}. The anthropometry variations (\ie the body shape) are ignored, which causes inconsistency between estimated local motions and global translations, as illustrated in Figure~\ref{fig:teaser} (b).
To address these challenges, we propose to lift the 2D human pose estimation to 3D perception by replacing the monocular camera with a camera of 3D capability.
In this paper, we use the stereo camera, as opposed to RGB-D cameras equipped with active IR or ToF sensors, for two reasons.
First, stereo cameras can be built on top of plain RGB cameras, making them potentially more cost-effective and available.
Second, we can make full use of those well-established monocular methods to make initial 2D predictions and then lift them to 3D coordinates directly based on the camera model.

In this paper, we propose the Stereo-Inertial Poser, which estimates \textit{metric-accurate} and \textit{shape-aware} human motions from six sparse IMUs and a single stereo camera.
We first introduce a stereo pose and body shape estimation module, making initial 3D estimations from the visual modality.
From the detected 3D keypoints and IMU measurements, we then use state space models to predict the intermediate 3D human joint positions and body root movements.
Finally, a shape-aware fusion module is used to fuse these intermediate results and the body shape parameters together, and the refinement network predicts local human motions and global translations consistent in metric 3D coordinates.
Our inference pipeline can achieve very high speed (more than 200 FPS) thanks to the end-to-end design, without the optimization-based post-processing procedures commonly used in previous works.
Extensive quantitative and qualitative evaluations demonstrate our method can produce accurate global translation and reduce the foot-skating effect compared with baseline methods.

Our main contributions can be summarized as:

(1) The Stereo-Inertial Poser, a metric-accurate human motion capture pipeline with a single stereo camera and six IMUs. It accurately estimates both local motions and full 3D trajectories in the metric coordinate system.
    
(2) The 3D human pose and shape estimation modules. The pose module lifts pose detections from well-established monocular methods to metric 3D coordinates. The shape module optimizes the initial body shape parameters.

(3) A shape-aware visual-inertial fusion module and related loss functions that enable the correct response to different body shapes without post-processing optimizations.

\section{Related Work}

The proposed method falls into the category of human motion capture by fusing a single camera and sparse IMUs.
It is also closely related to those methods that use either visual or inertial modality.

\subsection{Single Modal Human Motion Capture}

\subsubsection{Vision-Based Methods}
The gold standard for motion capture has historically been optical systems using reflective markers, which provide high precision but are confined to specialized studios~\cite{mocap_book}. To improve accessibility, research shifted towards markerless methods, first with multi-camera setups~\cite{mvcs_e2e,xu20234k4d} and more recently to systems using only a single monocular camera.
The advent of deep learning enabled motion capture from monocular videos. Early methods trained on large-scale 2D image datasets~\cite{coco,mpii} predicted 2D or pseudo-3D keypoints in the image coordinate system~\cite{conv_pose,crowdpose,openpose}. These methods lacked scale awareness and 3D spatial reasoning in world coordinates.
To infer 3D structure, researchers integrated parametric body models like SMPL~\cite{SMPL}. Early works optimized SMPL parameters against 2D keypoints using iterative fitting~\cite{smplify,smpl-x} or geometric constraints~\cite{tan_indirect_2017}. Later, direct regression networks~\cite{romp,Kocabas_PARE_2021,vibe} realized real-time inference, but they still remained limited to camera-relative 3D coordinates.
While it's possible to use structure-from-motion techniques to get the absolute 3D predictions, it only works with a moving camera, which is usually head-mounted~\cite{li2024coin,Kocabas2023pace,shen2024gvhmr}.
However, there is still need for finding metric 3D keypoints in world coordinates with a single fixed camera. This need motivated the use of cameras with 3D capabilities.
While RGB-D cameras are one option, the domain of 3D pose estimation from point clouds (\ie RGB-D images) is less mature than its 2D counterpart~\cite{rgbd2013,rgbd2021}.
Stereo cameras, however, present a powerful alternative. A key advantage of the stereo approach is that we can directly use those well-established 2D pose estimators to predict on each image in the stereo pair. The resulting 2D keypoints can then be triangulated into metric 3D coordinates using the known camera geometry~\cite{stereo_pose}.
This design choice strategically leverages the immense progress in the 2D domain to build a robust foundation for metric 3D perception, avoiding the need to develop a new 3D estimator from scratch.

\subsubsection{Inertial-Based Methods}
Current commercial solutions use dense IMUs (typically more than 15) to fully determine the orientations of all human parts and precise kinematics.
The research community focused on reducing the number of IMUs to make the system less intrusive, more convenient to set up, and more affordable.
Early methods started by focusing on the local body movements, ignoring the body's translation in space.
For example, SIP~\cite{sip} and DIP~\cite{DIP} proposed to recover human motions from 6 IMUs. This 6-IMU setting is followed by later works.
TransPose~\cite{transpose} is the first method to predict the body's global translation in open space, by utilizing the foot-ground contact hypothesis.
TIP~\cite{tip} and DynaIP~\cite{dynaip} adopt more advanced network structures to further improve precision.
PIP~\cite{pip}, PNP~\cite{pnp}, and GlobalPose~\cite{globalpose} progressively integrate advanced physics priors into the pipeline, making them work in complex settings and produce realistic motions.
Despite these advances, the drifting issue is inherent in inertial systems. Therefore, it's necessary to fuse visual and inertial modalities for precise metric root translation~\cite{robustcap}.

\subsection{Human Motion Capture with Visual-Inertial Fusion}

Fusing multi-view cameras with IMUs for human motion capture has been studied for a long time~\cite {zhang_fusing_2020,video_imu_pioneer}. These methods require cameras to be mounted in a panoptic-like way and well-calibrated.
Some works use a head-mounted camera and the underlying SLAM algorithm to capture the 6D pose of the human head~\cite{ego4o,mocap_every}.
However, fusing a single, fixed, third-person-view camera with sparse IMUs, is a different track.
In this track, early works~\cite{mono_imu_pioneer,hybridfusion,helten_2013} formulate it as an optimization problem to minimize energy functions of the visual and inertial signals.
In the era of deep learning, researchers focused on monocular fusion~\cite{hybridcap,chen2024}.
A recent work~\cite{chen2024} proposes a large-scale dataset and a baseline method to be available to the public.
The current state-of-the-art method, RobustCap~\cite{robustcap}, demonstrates impressive robustness but inherits the fundamental scale ambiguity of the monocular camera. While they can produce smooth and plausible trajectories, the absolute scale of the motion relative to the environment remains unknown.
In this paper, we argue and demonstrate that by introducing the stereo camera, our system achieves true metric accuracy and shape-aware motion estimations while maintaining the low-cost, portable setup that makes hybrid systems attractive.

\section{Method}

\begin{figure*}
    \centering
    \includegraphics[width=0.95\linewidth]{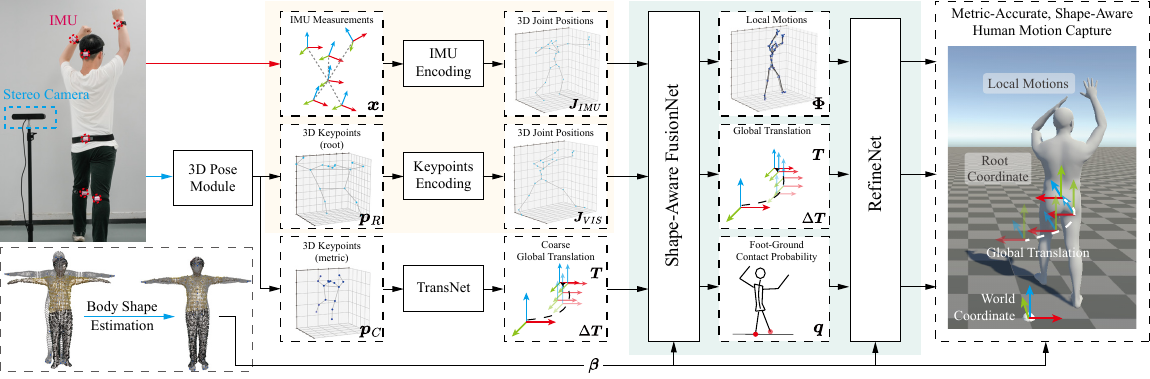}
    \caption{
    Our pipeline starts with one stereo camera and six IMUs.
    From the stereo image pairs, the 3D Pose Module predicts 3D keypoints in the root coordinate $\boldsymbol{p}_R$ and the world coordinate $\boldsymbol{p}_C$. The Body Pose Module estimates the body shape parameters $\beta$ (Sec.~\ref{sec:pose_shape}).
    These initial measurements and predictions are encoded by three separate networks (Sec.~\ref{sec:method_fusion}) into initial global translations $\boldsymbol{T}$, velocities $\Delta\boldsymbol{T}$, and 3D joint positions $\boldsymbol{J}_{IMU}$, $\boldsymbol{J}_{VIS}$ in the root coordinate.
    The FusionNet and RefineNet (Sec.~\ref{sec:method_fusion}) fuse these intermediate results and further refine them into metric-accurate, shape-aware final results, including local motions $\boldsymbol{\Phi}$, global translations $\boldsymbol{T}$, and the auxiliary contact probability $\boldsymbol{q}$.
    }
    \label{fig:pipeline}
\end{figure*}

\subsection{Overview}
\label{sec:method-overview}

Figure~\ref{fig:pipeline} shows the overview of the proposed pipeline,
which aims to recover human motion with six wearable IMUs and one external stereo camera.
The setup is in line with prior works~\cite{transpose,robustcap,pnp}, where the IMUs are mounted on the root joint (\ie the pelvis) and five leaf joints, including the head, forearms, and lower legs, while the third-person view camera is fixed.
We can get 3D human keypoint detections and body shape estimations from the stereo camera (Section~\ref{sec:pose_shape}).
The shape-aware visual-inertial fusion module fuses the visual detection results along with IMU signals, and finally predicts local motions represented by the SMPL model~\cite{SMPL} and global translations (Section~\ref{sec:method_fusion}).

\subsection{3D Pose and Body Shape Estimation}
\label{sec:pose_shape}

\subsubsection{3D Metric Pose Estimation}
\label{sec:3dkpts}

As mentioned in previous sections, a stereo camera, with known intrinsics and extrinsics, is the practical choice for fully using the well-established monocular pose estimation methods to obtain the metric 3D keypoints in the world coordinate.

At each timestamp $t$, the two synchronized frames from the left and right cameras, $F_{t,l}$ and $F_{t,r}$ are first processed by the widely-used MediaPipe Pose Landmarker toolbox~\cite{mediapipe}.
Then we get the pose detections, $\boldsymbol{p}_{2d,l}$, $\boldsymbol{p}_{2d,r}$, $\boldsymbol{p}_{3d,l}$, and $\boldsymbol{p}_{3d,r}$, respectively, where $\boldsymbol{p}_{2d}\in \mathbb{R}^{2N}$ are 2D keypoints on the image plane and $\boldsymbol{p}_{3d}\in \mathbb{R}^{3N}$ are canonical 3D predictions relative to the human body's root coordinate.
We use the layout of $N=17$ keypoints, as defined in the widely-used COCO dataset~\cite{coco}.
Each detected keypoint $p_i \in \boldsymbol{p}$ is associated with a scalar $c_i \in [0,1]$, representing how confident the model is, or, in other words, the visibility of that point.
We denote the confidence in vector form as $\boldsymbol{c}_{l}$ and $\boldsymbol{c}_{r} \in \mathbb{R}^{N}$.
Then, we can fuse the 3D keypoints estimated from the left and the right camera by the confidence,
\begin{equation}
    \boldsymbol{p}_{R} = \boldsymbol{\lambda}_c\boldsymbol{p}_{3d,l}    +    (1- \boldsymbol{\lambda}_c)\boldsymbol{p}_{3d,r}
    ,
\end{equation}
where $\boldsymbol{\lambda}_c = \boldsymbol{c}_{l} \cdot (\boldsymbol{c}_{l}+\boldsymbol{c}_{r})^{-1}$.
However, the 3D keypoints $\boldsymbol{p}_{R}$ are relative to the root coordinate of the human body. To acquire the metric body pose, we need to make full use of the stereo camera.
We suppose the stereo camera setup has a known baseline distance $d_{base}$ and shared intrinsics $\boldsymbol{K}$, with the world coordinate aligned with the left lens.
Then, with paired 2D keypoints $\boldsymbol{p}_{2d,l}$ and $\boldsymbol{p}_{2d,r}$, we derive the depth from each keypoint to the camera plane by
\begin{equation}
    d_{z} = f_x \frac{d_{base}}{d_{disp}}
    ,
\end{equation}
where $f_x\in \boldsymbol{K}$ is the horizontal focal length, $d_{disp} = | x_r - x_l |$ is the disparity for each point pair $p_l(x_l, y_l)\in \boldsymbol{p}_{2d,l}$ and $p_r(x_r, y_r)\in \boldsymbol{p}_{2d,r}$.
Then we have the following homogeneous equation
\begin{equation}
    \boldsymbol{p}_{C} = \boldsymbol{\lambda}_c \boldsymbol{d}_{z} \boldsymbol{K}^{-1}  \boldsymbol{p}_{2d,l} + 
    (1-\boldsymbol{\lambda}_c)  \left( \boldsymbol{d}_{z} \boldsymbol{K}^{-1}    \boldsymbol{p}_{2d,r} + \boldsymbol{t}_r \right)
    ,
\end{equation}
where $\boldsymbol{t}_r=[d_{base}, 0, 0]^T$ translates from the right lens to the world frame. The resulting 3D keypoints $\boldsymbol{p}_C$ in the world coordinate are fused by the detections according to the confidence.
The corresponding confidence $\boldsymbol{c}_C$ can be derived by averaging $\boldsymbol{c}_l$ and $\boldsymbol{c}_r$.
Now we have the 3D keypoints $\boldsymbol{p}_R$ in the root coordinate for further estimating the local motions, and metric 3D keypoints $\boldsymbol{p}_C$ in the world coordinate for the global translation.

\subsubsection{Body Shape Estimation}
\label{sec:body_shape_est}
Body shape is important for consistency between the local motions and global translations.
Therefore, we need to estimate each subject's body shape by fitting the SMPL model into the subject.
We first ask the subject to stand in front of the camera and keep it in T-pose, which aligns with the rest pose of SMPL.
We then have the raw point cloud $\mathcal{P}_{raw}$ by stereo matching~\cite{OpenStereo}, as well as the 3D skeleton $J_{raw} = \boldsymbol{p}_C$ following the previous section.
The point cloud is then segmented according to the subject's bounding box, and further voxel-downsampled into around $4000$ points $\mathcal{P}_{down}$ for computational efficiency.
Besides, a 6D transformation, denoted by matrix $\boldsymbol{M} = [\boldsymbol{R}|\boldsymbol{t}]$, should be applied to $\mathcal{P}_{down}$ and $J_{raw}$ in order to align the coordinate between the camera and SMPL. We denote the transformed point cloud as $\mathcal{P}$, and the transformed skeleton as $J$.
The body shape estimation is formulated as the minimization of a composite energy function:
\begin{equation}
    \boldsymbol{E}(\boldsymbol{\beta},\boldsymbol{\Phi},\boldsymbol{M})
    = \lambda_{\mathrm{skel}}\,E_{\mathrm{skel}}
    + \lambda_{\mathrm{cd}}\,E_{\mathrm{cd}}
    + \lambda_{\Phi}\,E_{\Phi}
    + \lambda_{\beta}\,E_{\beta}
    ,
\end{equation}
where the body shape parameters $\boldsymbol{\beta}$ and the body pose parameters $\boldsymbol{\Phi}$ are from the SMPL model.
The skeletal alignment
\begin{equation}
    E_{\mathrm{skel}}
    = \sum_{i=1}^{N} \bigl\lVert J_i - \hat J_i\bigr\rVert^2
\end{equation}
penalizes the distance between observed 3D skeleton $J$ and fitted joints $\hat J = f(\boldsymbol{\beta}, \boldsymbol{\Phi})$, where $f(\cdot)$ is the SMPL forward kinematics function.
The Chamfer distance
\begin{equation}
    E_{\mathrm{cd}}
    = \frac{1}{|\mathcal P|}\sum_{p\in \mathcal P}\min_{v\in V}\|p-v\|^2
      + \frac{1}{|V|}\sum_{v\in V}\min_{p\in \mathcal P}\|v-p\|^2
\end{equation}
describes the distance between the point cloud $\mathcal P$ and the SMPL mesh vertices $V$ constructed by $\boldsymbol{\Phi}$ and $\boldsymbol{\beta}$.
The pose prior
$
    E_{\Phi}
    = \|\boldsymbol{\Phi}\|^2
$
and the shape prior
$
    E_{\beta}
    = \|\boldsymbol{\beta}\|^2
$
regularize the pose parameters $\boldsymbol{\Phi}$ and shape parameters $\boldsymbol{\beta}$ towards $\boldsymbol{0}$.
We empirically set $\lambda_{\mathrm{skel}}=1$, $\lambda_{\mathrm{cd}}=15$, $\lambda_{\Phi}=1$, and $\lambda_{\beta}=0.01$. The energy function is solved by an SGD with momentum optimizer in a few seconds.
In the supplementary, we visualize the results of four subjects.

\subsection{Shape-Aware Visual-Inertial Fusion}
\label{sec:method_fusion}

Full human motions in 3D space can be decoupled into two elements: global translations and local motions.
In line with prior work~\cite{robustcap}, the global translation is defined as the translation $\boldsymbol{T}_{Cr}$ of the root pelvis joint in the world coordinate. The local motion is defined as the SMPL joint pose parameters $\boldsymbol{\Phi} \in \mathbb{R}^{24\times 3}$.
This section describes estimating metric-accurate and shape-aware human motions from IMU measurements and detected 3D keypoints, involving the initial global translation and local motion estimation, and the final shape-aware visual-inertial fusion.

\subsubsection{Initial Global Translation Estimation}
\label{sec:global_t_est}
We build the TransNet with the State Space Model (SSM)~\cite{mamba} building block
with MLP layers
to learn the inherent temporal-spatial features in the 3D keypoints and predict the global translation.
The global translation is more directly related to the keypoints of the head and the body trunk than the limbs.
Therefore, we select a subset of $9$ keypoints $\boldsymbol{p}_C'$ from $\boldsymbol{p}_C$ as the network inputs, including the nose, eyes, ears, shoulders, and hips.
To help the networks learn from the high-frequency signals better,
the coordinates are linearly scaled to $[0, 1]$ and further processed by the positional encoding~\cite{nerf} with width $d=4$.
The encoded coordinates and the corresponding confidence $\boldsymbol{c}_{C}'$ are concatenated as the network inputs $\boldsymbol{x}\in \mathbb{R}^{n}$, where $n=9\times (8\times 3+1)$.
The TransNet predicts the global translation vector in the metric camera coordinate $\boldsymbol{T} \in \mathbb{R}^3$
and also the change of translation between two frames $\Delta\boldsymbol{T} \in \mathbb{R}^3$.
We use $L_2$ loss to supervise $\boldsymbol{T}$,
\begin{equation}
    \mathcal{L}_{T} = ||\boldsymbol{T} - \boldsymbol{T}^{GT}||_2^2 .
\end{equation}
We apply a spatial-temporal cycle consistency loss to supervise $\Delta\boldsymbol{T}_{t}$,
\begin{equation}
    \mathcal{L}_{\Delta T_{t}} = || \Delta\boldsymbol{T}_{t} -(\boldsymbol{T}_{t} - \boldsymbol{T}_{t-1}) ||_2^2 + || \Delta\boldsymbol{T}_{t} - \Delta\boldsymbol{T}_{t}^{GT}||_2^2 .
\end{equation}
Therefore, the final loss for the global pose estimation network goes $\mathcal{L}_T = \sum_t \left( \mathcal{L}_{T_{t}} + \mathcal{L}_{\Delta T_{t}} \right)$.
Please note that the proposed method does not rely on specific network structures.
We use the SSM mainly because it balances accuracy and performance well. It's possible to use other temporal models like plain RNNs~\cite{rnn}, LSTMs~\cite{lstm}, and Transformers~\cite{transformer}.

\subsubsection{Initial Local Motion Estimation}
\label{sec:local_motion_est}
Directly estimating human motions from two different modalities is a difficult problem. Therefore, we use two separate networks, the IMU Encoding Network (IENet) and the Keypoints Encoding Network (KENet), to first unify these two modalities into the canonical root joint space, which is defined by the forward kinematics function from the SMPL model.

\paragraph{IMU Encoding Network}
We use an SSM to predict the 3D joint positions from the IMU measurements. The linear accelerations $\boldsymbol{\alpha}\in \mathbb{R}^{6\times 3}$ and rotations $\boldsymbol{R}^{6\times 3 \times 3}$ are first transformed into the root frame's coordinate, according to the rotation measured by the IMU mounted on the pelvis.
The linear accelerations are again processed by positional encoding, and the rotations are converted to the 6D vector representation~\cite{continuty_rot}. These pre-processing techniques can help the network learn better. Then the accelerations $\hat{\boldsymbol{\alpha}}\in \mathbb{R}^{6\times 3\times 8}$ and rotations $\hat{\boldsymbol{R}}\in \mathbb{R}^{6 \times 6}$ are flattened and concatenated together into the input vector $\boldsymbol{x}_t \in \mathbb{R}^{180}$. The network continuously takes in a sequence of $\boldsymbol{x}_t$ and predicts the 3D joint positions of the SMPL model $\boldsymbol{J}_{IMU} \in \mathbb{R}^{24\times 3}$ in the canonical root coordinate.
We use $L_2$ loss to supervise the network,
\begin{equation}
    \mathcal{L}_{J_{IMU}} = ||\boldsymbol{J}_{IMU} - \boldsymbol{J}^{GT}||_2^2 .
\end{equation}

\paragraph{Keypoints Encoding Network}
In parallel to the IENet, another SSM is used to predict 3D joint positions from the 3D keypoints $\boldsymbol{p}_R$ and confidence $\boldsymbol{c}_C$.
Given a camera frame, $\boldsymbol{p}_R$ is first normalized into $[0,1]$ according to the minimum and maximum values of each axis.
The normalized $\hat{\boldsymbol{p}}_R$ and $\boldsymbol{c}_C$ are then positional-encoded similarly. The KENet predicts the 3D joint positions $\boldsymbol{J}_{VIS}$ in the canonical root coordinate. We use $L_2$ loss to supervise it,
\begin{equation}
    \mathcal{L}_{J_{VIS}} = ||\boldsymbol{J}_{VIS} - \boldsymbol{J}^{GT}||_2^2 .
\end{equation}
The KENet is important, since the input 3D keypoints are defined slightly differently from the output 3D joint positions.
Also, it's good for adaptability and compatibility. Since different 3D keypoint estimation models may use different definitions of keypoints, such a modular design makes it possible to switch to new state-of-the-art pose estimation methods easily, avoiding re-training the whole pipeline.

\subsubsection{Shape-Aware Visual-Inertial Fusion}
\label{sec:fusion_module}
Now, we can fuse the intermediate results and raw measurements together using FusionNet to get our desired output.
The inputs include the canonical joint positions $\boldsymbol{J}_{IMU}$ and $\boldsymbol{J}_{VIS}$,
the confidence $\boldsymbol{c}_C$,
the body shape parameters $\boldsymbol{\beta}$,
the initial translations $\boldsymbol{T}$ and $\Delta \boldsymbol{T}$,
and the IMU measurements $\boldsymbol{x}$.
We apply positional encoding on $\boldsymbol{J}_{IMU}$, $\boldsymbol{J}_{VIS}$, $\boldsymbol{T}$, $\Delta \boldsymbol{T}$ and $\boldsymbol{x}$.
The FusionNet is shape-aware, and it predicts the refined global translation $\boldsymbol{T}$ and $\Delta\boldsymbol{T}\in \mathbb{R}^{3}$, and joint rotations in axis-angle representation $\boldsymbol{\Phi} \in \mathbb{R}^{24\times 3}$ for local motions.
We use the $L_2$ loss $\mathcal{L}_T$ and the cycle consistency loss $\mathcal{L}_{\Delta T}$ to supervise the refined global translation, as described in Section~\ref{sec:global_t_est}.
The loss function for supervising $\boldsymbol{\Phi}$ is defined as
\begin{equation}
\begin{split}
    \mathcal{L}_{\Phi} & = \mathcal{L}_{rot} + \lambda \mathcal{L}_{fk} \\
                       & = ||\boldsymbol{\Phi} - \boldsymbol{\Phi}^{GT}||^2_2 + \lambda  || f(\boldsymbol{\Phi}, \boldsymbol{\beta}) - f(\boldsymbol{\Phi}^{GT}, \boldsymbol{\beta}) ||_2^2 ,
\end{split}
\end{equation}
where $\lambda$ is a scalar for balancing the weights. $f(\cdot)$ is a kinematics function that takes in joint rotations and body shape, and outputs the 3D positions of each joint. We empirically set $\lambda = 2.5$.

Additionally, we design some auxiliary outputs and loss functions to help train the network.
To prevent the foot-skating effect, we introduce the foot-ground contact vector $\boldsymbol{q} = <q_l, q_r> \in [0,1]^2$ to indicate if the left or the right foot stably contacts the ground. The binary cross-entropy loss is applied to supervise it,
\begin{equation}
    \mathcal{L}_{fc} = - \left(  \boldsymbol{q}^{GT}  \log \boldsymbol{q}
         + \left(  (\boldsymbol{1}-\boldsymbol{q}^{GT}  \right) 
         \log \left(  \boldsymbol{1}-\boldsymbol{q}  \right)  \right)
         .
\end{equation}
Then, we define the foot-skating loss at given timestamp $t$,
\begin{equation}
    \mathcal{L}_{fs, t} = \sum_{j\in\{l,r\}} q_j || f_{j}(\boldsymbol{\Phi}_t, \boldsymbol{\beta}) - f_{j}(\boldsymbol{\Phi}_{t-1}, \boldsymbol{\beta}) + \Delta \boldsymbol{T}_t ||_2^2
    ,
\end{equation}
where $f_{j}(\cdot)$ indicates the forward kinematics function that maps joint rotations and body shape into the 3D position of a given joint $j$ in the root coordinate, and $j$ indicates the left or the right foot.
It can be intuitively interpreted as, if the foot contact does not slide with the ground, then its movements in the root coordinate should exactly reverse the body root's movement in the world coordinate.
Similarly, we also define a jerk loss on each joint position at time $t$ to stabilize the output motion,
\begin{equation}
    \mathcal{L}_{jk,t} = \sum_j || f_{j,t} - 3f_{j,t-1} + 3 f_{j,t-2} - f_{j,t-3} ||^2_2
    .
\end{equation}

Finally, we add those loss functions together,
\begin{equation}
    \mathcal{L} = \lambda_{\Phi}\mathcal{L}_{\Phi}
                         + \lambda_{T}\mathcal{L}_{T}
                         + \lambda_{\Delta T}\mathcal{L}_{\Delta T}
                         + \lambda_{fc}\mathcal{L}_{fc}
                         + \lambda_{fs}\mathcal{L}_{fs}
                         + \lambda_{jk}\mathcal{L}_{jk}
                         .
\end{equation}
We empirically set $\lambda_{\Phi} = 20$, $\lambda_{T} = 5$, $\lambda_{\Delta T} = 5$, $\lambda_{fc} = 0.001$, $\lambda_{fs} = 100$, and $\lambda_{jk} = 50$.

\subsubsection{Final Refinement Network}
\label{sec:refine_module}
In practice, we find it helpful to add an extra refinement module, RefineNet, to refine the coarse results of the fusion module.
The refinement module shares a similar design with previous modules.
The input includes the FusionNet predictions, including local motions $\boldsymbol{\Phi}$, global translations $\boldsymbol{T}$, velocities $\Delta\boldsymbol{T}$, and foot-ground contact probability $\boldsymbol{q}$, along with the body shape $\beta$.
The output and loss function design are the same as the fusion module.
We conduct ablation studies on the effectiveness of RefineNet in Section~\ref{sec:ablation}.

\section{Experiments}

\begin{table*}
\caption{Quantitative comparison with other methods on the AIST++ and TotalCapture dataset.}
\label{tab:exp}
\centering
\small
\begin{tabular}{ccccccccccccc}
  \toprule
        & \multicolumn{6}{c}{AIST++}& \multicolumn{6}{c}{TotalCapture}\\
          \cmidrule(r{1ex}){2-7}          \cmidrule(r{1ex}){8-13}
  Method& JPE &PVE &SIP    &TE &Jerk &FS           & JPE & PVE & SIP & TE & Jerk &FS \\
\midrule
  PIP~\cite{pip}  &87.0&116.5&28.1   & 45.2 &\textbf{1.04} &0.91        &49.1 &66.0 & 12.9& 43.8 & \textbf{0.20}& \textbf{0.43}\\
  PNP~\cite{pnp}  & -& -& -& -& -& -                          &47.4 &- &    10.8 &- &0.26 &-\\
  GlobalPose~\cite{globalpose} &- &- &- &- &- &-                     &\textbf{43.1} &\textbf{49.6} &   - &- & 0.21 &-\\
  HybridCap~\cite{hybridcap}& 33.3 & -& -&- &- & -&- &- &- &- &- &-\\
  RobustCap~\cite{robustcap} & 33.1&43.2&9.34&  9.91 & 2.49 &0.61   &48.7 & 63.4&13.4&  23.5 & 2.09 &1.4\\
\midrule
  Ours \textit{(ideal)}               &\textbf{29.8}&\textbf{39.0}&\textbf{8.91}&\textbf{1.13}&1.50&\textbf{0.57}    &46.5&54.3&\textbf{10.6}   &\textbf{5.42}&0.94&0.69\\
  Ours \textit{(noise lvl. 5)}        &30.6&39.8&9.12&4.83&1.95&0.58    &47.6&55.8&11.0   &8.93&1.39&0.71\\
  Ours \textit{(noise lvl. 15)}       &35.5&46.8&10.1&13.2&2.45&0.63    &51.1&60.1&11.4   &15.8&1.92&0.79\\
  Ours \textit{(syn. stereo)}         &31.7&43.1&9.89&7.76&1.74&0.60    &49.0&57.1&10.9   &9.41&1.73&0.72\\
  RobustCap3D \textit{(ideal)}        &33.0&43.2&9.27&6.97&2.34&0.57    &48.0&60.8&12.0   &14.1&1.35&0.71\\
  RobustCap3D \textit{(noise lvl. 5)} &35.6&43.4&9.35&7.58&2.74&0.60    &49.1&61.6&12.5   &16.4&1.46&0.75\\
  RobustCap3D \textit{(noise lvl. 15)}&38.1&48.9&10.6&16.8&4.00&0.75    &52.9&66.5&13.3   &21.9&2.01&0.99\\
  RobustCap3D \textit{(syn. stereo)}  &36.0&46.1&11.8&9.91&3.31&0.60    &50.8&63.2&12.9   &16.9&1.68&0.81\\
  \bottomrule
\end{tabular}
\end{table*}

\begin{table}
\caption{Ablation studies on the AIST++ dataset.}
\label{tab:abl}
\centering
\small
\begin{tabular}{lcccccc}
  \toprule
  Method& MPJPE &PVE &SIP &TE &Jerk &FS\\ \midrule
  Ours& 29.8& \textbf{39.0}& 8.91& \textbf{1.13}& \textbf{1.50}&\textbf{0.57}\\
  - cano. space& 30.3& 39.7& 8.91& \textbf{1.13}& 1.58&\textbf{0.57}\\
  - shape-aware & 31.1& 40.8& 8.81& 1.26& 1.57&0.64\\
  - pos. encoding & 29.9& 39.1& 8.82& 1.60& 1.65&0.60\\
  - RefineNet& 31.5& 41.4& 8.90& 1.36& 2.15&0.63\\
  - jerk loss& 30.2& 39.9& \textbf{8.80}& 1.49& 4.56&0.79\\
  - cycle loss& \textbf{29.7}& 39.0& 9.09& 1.38& 1.53&0.58\\
  - foot skt. loss& 30.4& 39.8& 8.91& 2.84& 1.65&0.66\\
  \bottomrule
\end{tabular}
\end{table}

\subsection{Implementation Details}

We run the training and evaluation processes on a desktop computer with an i7-14700 CPU, an RTX 4070 Ti Super GPU, and 64 GB RAM. We use the ZED 2 stereo camera, which features 720P resolution, 60 Hz rate, and a 120 degree diagonal field of view. The wearable IMUs are developed in-house within our laboratory, featuring high-performance IMU chips and precise synchronization mechanism. Since our method does not rely on specific stereo cameras or IMUs, users can adopt any other commercially-available cameras or IMUs. For more details, please refer to the supplementary.

\subsubsection{Datasets}

In line with prior work~\cite{robustcap},
we train our method on the AMASS dataset~\cite{amass} and the AIST++ dataset~\cite{aistpp}, with synthesized keypoints, IMUs, and the ground truth for foot-ground contact.
On AIST++, since it's built with multi-view RGB cameras, we use MediaPipe to generate the detections $\boldsymbol{p}_R$ in the root coordinate from each camera view and synthesize 3D keypoints $\boldsymbol{p}_C$ in the world coordinate by adding a random noise $\sigma = 5 \text{cm}$ on the ground truth value.
We follow the body shape parameters provided by the AMASS dataset.
We evaluate our method on the AIST++ dataset and the TotalCapture dataset~\cite{totalcapture}.
On these two datasets, for lack of ground truth shape labels, we use the proposed method (Section~\ref{sec:body_shape_est}) to estimate the body shape for each subject, with manually selected frames where the subject acts like the T-pose. Also, we ignore the chamfer distance item by setting $\lambda_{cd}=0$ for lack of point clouds.

\subsubsection{Metrics}
We evaluate our method on 6 metrics.
Joint Position Error (JPE) measures the mean Euclidean distance error of all joints in the root coordinate in mm.
PVE measures the per-vertex error in mm.
SIP Error measures the mean global rotation error of hips and shoulders in degrees.
Translation Error (TE) measures the absolute translational error between predicted root joint and the ground truth, in cm.
Jerk Error measures the averaged jerk (\ie the first time derivative of acceleration) in $km/s^3$.
Foot-Skating Error (FS) measures the averaged predicted foot movements in mm in the global space, when ground truth foot movement equals zero (\ie when $q_l$ or $q_r=1$).
All metrics follow the less is better principle.
The first three metrics measure the accuracy of local motions, while the latter three metrics are more related to global translations.

\subsection{Evaluations}

\textbf{Quantitative Results.}
We compare our method with the SOTA visual-inertial fusion methods HybridCap~\cite{hybridcap} and RobustCap~\cite{robustcap}, and SOTA inertial-only methods PIP~\cite{pip}, PNP~\cite{pnp}, and GlobalPose~\cite{globalpose}.
For fair comparison, we enhance and retrain RobustCap by replacing its 2D keypoints inputs with 3D keypoints, denoted by RobustCap3D, which should improve its global translation estimation.

The quantitative results are shown in Table~\ref{tab:exp}. Please note that some numbers are missing because they are not reported by the original authors.
During the evaluation, we adopt two different ways to synthesize 3D keypoints $\boldsymbol{p}_C$, including adding random gaussian noises to the ground-truth 3D keypoints, and building a synthetic, virtual stereo camera. We add random noises with three different levels $\sigma=0$ (denoted as \textit{ideal}), $5$ and $15$ cm.
Please note that the rated depth accuracy of common RGB-D cameras are usually around several millimeters to centimeters.
The synthetic stereo camera, denoted as \textit{syn. stereo}, is built by pairing adjacent RGB cameras from the dataset to simulate stereo geometry for detecting 3D keypoints $\boldsymbol{p}_C$.
Our method shows significant improvements in TE over baseline methods, except under the impractical 15 cm noise, which is way beyond typical depth noise ranges (\ie less than 10 centimeters).
On the AIST++ dataset with aggressive dance motions (\eg jumps, rolls), our method achieves SOTA performance across most metrics, attributed to the shape-aware fusion module that fully exploits metric-accurate 3D keypoints. Notably, RobustCap3D fails to exploit 3D keypoint advantages even with ideal inputs, as evidenced by its suboptimal TE. 
On TotalCapture, visual-inertial fusion methods underperform recent inertial-only baselines due to limited translational movements and some out-of-view frames with severe occlusion.
Nevertheless, our method outperforms all existing fusion approaches.
Finally, while recent inertial methods exhibit strong Jerk and FS metrics through physics-based post-optimization, their body shape-agnostic design may limit translational accuracy as illustrated in Figure~\ref{fig:teaser} but related metrics are not reported by the authors.

In terms of inference speed, in offline evaluation, our method runs at more than 200 FPS, thanks to the end-to-end structure without the need of post-processing. In the supplementary video, we also demonstrate online evaluations where our method is running in real time, synchronized with the camera at 60 FPS by the timestamps.

\textbf{Qualitative Results.}
We conduct extensive real-world experiments to evaluate our method against the baselines. 
The evaluations took place in a 6m $\times$ 6m motion capture field to assess long-term drift and accuracy of local motions. In a key experiment, a subject walked along a circular path with radius of 2 meters for 10 laps. PIP and PNP exhibit severe drift, especially during dynamic motions like jumping. RobustCap remains drift-free but produces a geometrically distorted trajectory, a consequence of monocular scale ambiguity. Our method produces both a drift-free and a metrically accurate trajectory, correctly recovering the circular path's scale and shape. 
For more detailed comparisons and the live demo, please refer to the supplementary video.

\subsection{Ablation Studies}
\label{sec:ablation}
We perform ablation studies on the AIST++ dataset with ideal $\boldsymbol{p}_C$.
Table~\ref{tab:abl} shows the results of our ablation studies, where we quantify the contribution of each individual component in our pipeline.
We first examine the design of the canonical root space for intermediate representations. Specifically, we do not normalize the 3D keypoints per frame as the KENet inputs, and make the KENet and IENet predict joint positions in the metric coordinate. The dropped performance in local motions could be attributed to different body shapes inherent in keypoints and joint positions which confuse the neural networks.
Next, we examine the shape-aware fusion module. By setting the input $\boldsymbol{\beta}=\boldsymbol{0}$ as the input during the training and the inference process, the fusion module is then unaware of the body shape. The resulting performance degradation in all metrics demonstrates the importance of shape-awareness for motion capture.
Then, we remove the positional encoding process and witness a rise in translation error and jerk, indicating the resulted networks fail to learn stably and thus proving the effectiveness of the positional encoding.
After that, we test the performance without the RefineNet by directly evaluating the initial outputs of the fusion module. The performance degradations in both local motions and global translations prove the refine module can bridge the gap between the local and the world coordinate.
Finally, we check the effectiveness of the loss function design, including the jerk loss, the foot-skating loss and the cycle consistency loss
, which suggests we can resolve the \textit{global translation} v.s. \textit{local motions} (\eg foot-skating) dilemma illustrated in Figure~\ref{fig:teaser} (b) by introducing correct shape-awareness and proper supervision.

\subsection{Limitations}

In our pipeline, despite the IMUs supporting up to 400 Hz sampling rate, our system synchronizes them with the camera at 60 FPS. Future work could decouple the temporal visual-inertial fusion in an asynchronous manner to fully exploit the high-frequency inertial measurements for transient motion estimation.
Second, our shape-aware fusion is the initial step towards improving anthropometric modeling in this track through implicit neural networks. Future work could focus on making comprehensive use of the body shape, including bone-length and soft-tissue variations (\eg body fat or muscle mass).
Third, akin to many existing approaches, we utilize IMU acceleration and orientation but under-utilize gyroscope data, which could be further utilized with physics-based constraints.
Finally, the current stereo camera setup imposes limits on the workspace area and may be sensitive to poor lighting conditions.

\section{Conclusion}
We presented Stereo-Inertial Poser, a real-time motion capture system that integrates a stereo camera with six IMUs. Our method resolves depth ambiguity through calibrated stereo geometry while compensating for inertial drift via state-space fusion. The novel shape-aware fusion module ensures anthropometric consistency between global translations and local motions. Experiments demonstrate our system achieves significant improvements in global translation accuracy and reduced foot-skating artifacts compared to state-of-the-art baselines. With real-time inference performance, our approach is practical for applications in robotics. Our work bridges the gap between cost-effective hardware and professional-grade motion fidelity, advancing democratized tools for human motion capture and human-robot interaction.

\bibliographystyle{IEEEtran}
\bibliography{bibtex}

\end{document}